\documentclass[conference]{IEEEtran}
\IEEEoverridecommandlockouts
\usepackage{cite}
\usepackage{amsmath,amssymb,amsfonts}
\usepackage{algorithmic}
\usepackage{graphicx}
\usepackage{textcomp}
\usepackage{xcolor}
\def\BibTeX{{\rm B\kern-.05em{\sc i\kern-.025em b}\kern-.08em
    T\kern-.1667em\lower.7ex\hbox{E}\kern-.125emX}}
    
\usepackage{caption}
\usepackage{subcaption}

\usepackage{fancyhdr}

\begin{document}

%\title{Automated Embedding Density Based Clustering of Multiplex Cell Images\\
\title{Fast Data Driven Estimation of Cluster Number in Multiplex Images using Embedded Density Outliers\\
}

\author{\IEEEauthorblockN{Spencer Angus Thomas\textsuperscript{1,2}}
\IEEEauthorblockA{\textsuperscript{1}\textit{Dept. of Computer Science, University of Surrey, UK}\\
\textit{\textsuperscript{2}Data Science, National Physical Laboratory (NPL), UK}\\
s.a.thomas@surrey.ac.uk \\
0000-0002-9260-1561}
}

\maketitle
%copyright notice
\thispagestyle{plain}
\fancypagestyle{plain}{
\fancyhf{} % clear all header and footer fields
\fancyfoot[L]{978-1-6654-8462-6/22/\$31.00~\copyright 2022 Crown} % change copyright notice here if required
\renewcommand{\headrulewidth}{0pt}
\renewcommand{\footrulewidth}{0pt}
}

\begin{abstract}
The usage of chemical imaging technologies is becoming a routine accompaniment to traditional methods in pathology. Significant technological advances have developed these next generation techniques to provide rich, spatially resolved, multidimensional chemical images. The rise of digital pathology has significantly enhanced the synergy of these imaging modalities with optical microscopy and immunohistochemistry, enhancing our understanding of the biological mechanisms and progression of diseases. Techniques such as imaging mass cytometry provide labelled multidimensional (multiplex) images of specific components used in conjunction with digital pathology techniques. These powerful techniques generate a wealth of high dimensional data that create significant challenges in data analysis. Unsupervised methods such as clustering are an attractive way to analyse these data, however, they require the selection of parameters such as the number of clusters. Here we propose a methodology to estimate the number of clusters in an automatic data-driven manner using a deep sparse autoencoder to embed the data into a lower dimensional space. We compute the density of regions in the embedded space, the majority of which are empty, enabling the high density regions (i.e. clusters) to be detected as outliers and provide an estimate for the number of clusters. This framework provides a fully unsupervised and data-driven method to analyse multidimensional data. In this work we demonstrate our method using 45 multiplex imaging mass cytometry datasets. Moreover, our model is trained using only one of the datasets and the learned embedding is applied to the remaining 44 images providing an efficient process for data analysis. Finally, we demonstrate the high computational efficiency of our method which is two orders of magnitude faster than estimating via computing the sum squared distances as a function of cluster number.
\end{abstract}


\section{Introduction}

Immunohistochemistry (IHC) is routinely used in the diagnostics of tissue pathology as it can visualise the expression of proteins by tagging with an enzymes or fluorophores to change the colour pigment of the tissue \cite{angelo_multiplexed_2014}.  
Multiplex IHC uses multiple types of stain in order to visualise many different components in tissues simultaneously.
Multiplex imaging technologies provide a powerful way to image tissue sections and visualise the spatial organisation of different cell types and differences between them \cite{baars_matisse_2021}. This mapping capability is vitally important in biological studies such as in oncology and cancer research by providing information about tissue heterogeneity and tumour micro-environment \cite{binnewies_understanding_2018}. 
However, when used in parallel, these tags can exhibit spatial and spectral overlap limiting their clinical usage \cite{angelo_multiplexed_2014}.

Another form of multiplex imaging is mass spectrometry based, such as  imaging mass cytometry (IMC) that uses metal conjugated antibodies to label and measure specific protein markers in tissues \cite{giesen_highly_2014}. These are used in a spatially resolved manner to provide multidimensional images where each pixel corresponds to the measured labels at sub cellular resolution. 
The rich chemical information in IMC enhances the study of tissue revealing tumour heterogeneity, cell-cell interactions, and moving towards individualised diagnosis and therapies \cite{giesen_highly_2014}.

A significant challenge for this data is the segmentation of individuals cells in the multiplex images. %A number of pipelines have been proposed
%Image segmentation and clustering for high dimensional data is a long standing challenge. 
The analysis of high dimensional data can be aided by dimensionality reduction, for example two or three dimensional projections can reveal visual patterns in the low dimensional space. Patterns such as manifolds or clusters in the data can provide insight into structure in images \cite{Maaten2008}, similarity of text documents \cite{Hinton2006, LeCun2015}, informative overviews of hyperspectral data \cite{Fonville2013}, and sub-groups of genes \cite{Mahfouz2015} or cell expression \cite{Macosko2015}. This provides a powerful computational tool to efficiently analyse the chemical information in mass spectromertry data, such as IMC \cite{abdelmoula_data-driven_2016}. Methods such as t-distributed stochastic neighbour embedding (t-SNE) \cite{Maaten2008} are state of the art techniques data reduction and visualisation. However the lack of a known mapping prohibits the application to unseen data \cite{dexter_training_2020}. Autoencoders avoid this issue by learning the encoding and decoding transformation during training of the model \cite{Thomas2019CMR}. Despite their ability to learn low dimensional representations of high dimensional data, both t-SNE and autoencoders do not segment the learned patterns directly and require additional methods or training phases. Clustering data introduces additional challenges in the estimation of the number of clusters, which must either been known \emph{a priori} or optimised in expensive computations. Density based clustering methods such as DBSCAN \cite{ester_density-based_1996} do not require a number of clusters, though require selection of a minimum number of points and radius parameter, which may not be intuitive or easy to optimise. 

For multiplex data such as IMC, computational approaches lack the ability to fully explore the rich spatially resolved multiplexed (high dimensional) tissue measurements \cite{schapiro_histocat_2017}. Therefore in this work we develop a method to efficiently analyse these data in a purely data-driven way. Our method first embeds the data into a (reduced) three dimensional space, which we interpret as pseudo red blue green (RGB) components in providing a rich single image summary of the data. Next, we compute the density of points in binned regions of the embedded space. Finally we use the number of dense regions as an estimate for the number of clusters in the data and use this number to perform $k$-means clustering of the embedded space. The use of an autoencoder enables unseen data to be embedded as the transformations are known, and the density estimate for the number of clusters allow data specific clustering. We demonstrate this method by training our model on one IMC dataset and applying this to 44 unseen multiplex images. 

\section{Data}

\subsection{Multiplex Imaging}%\label{AA}
The data used in this work are multiplex images of thin tissue sections of human patients with inflammatory bowel disease obtained from \cite{baars_matisse_2021_data, baars_matisse_2021}. The imaging data contain multiple measurements for each pixel in the image, with each measurement corresponding to a different feature (more details are in Section \ref{sec:IMC}). This type of imaging is comparable to hyperspectral images where each pixel has multiple spectral components. 

\begin{figure*}[t]
    \centering
    \includegraphics[width=0.95\textwidth]{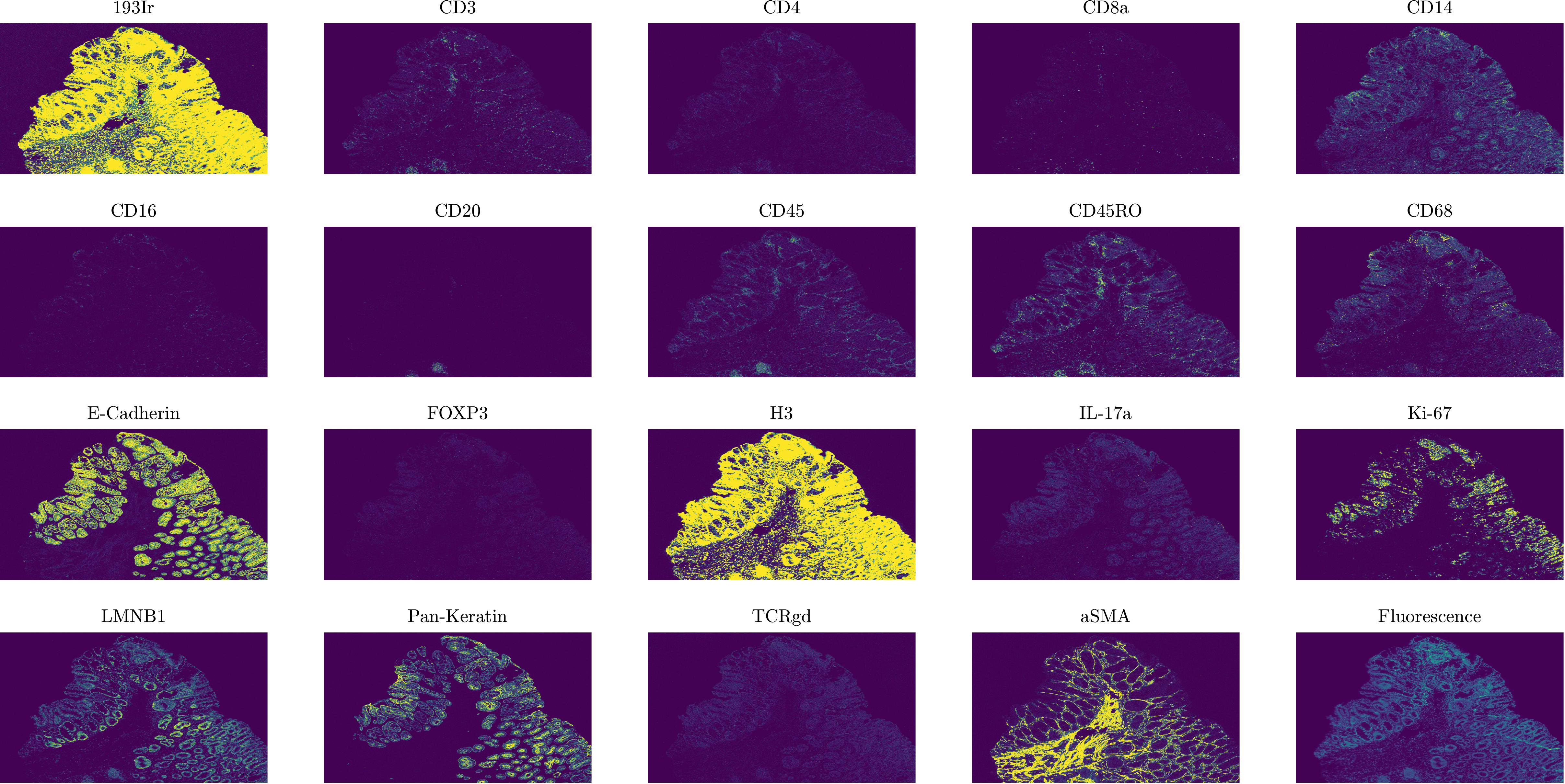}
    \caption{IMC feature maps for the training data in this work. Spatial distributions of the 19 markers in the IMC hyperspectral data (saturated for clarity of image) and corresponding Fluorescence image. }
    \label{fig:IMC}
\end{figure*}

\subsection{Imaging Mass Cytometry}\label{sec:IMC}
The images are obtained from formalin-fixed paraffin embedded colonic tissue biopsies from patients collected from \cite{baars_matisse_2021_data, baars_matisse_2021} with appropriate ethics. The data are acquired using Imaging Mass Cytometry (IMC) where the sample is tagged with labeled antibodies that target particular cellular proteins and a mass spectrometer is used to detect these tags at each pixel. The output is an image dataset with $N$ measured components for each $X$ and $y$ pixel. Each hyperspectral image consists of 19 labelled measurements for each pixel, see Fig.~\ref{fig:IMC},  %The ordered feature maps correspond to the spatial distribution of 193Ir, CD14, CD16, CD20, CD3, CD4, CD45, CD45RO, CD68, CD8a, E-Cadherin, FOXP3, H3, IL-17a, Ki-67, LMNB1, Pan-Keratin, TCRgd and aSMA
(for full details refer to \cite{baars_matisse_2021}).
%using the the antibodies listed in Table.~\ref{tab:labels}.
% t
% \begin{table}[]
%     \centering
%     \caption{Antibody labels used in the IMC data acquisition in \cite{baars_matisse_2021_data, baars_matisse_2021}}
%     \begin{tabular}{l|l}
%     \hline 
%         Name & Role \\
%     \hline 
% 193Ir	& DNA intercalator		\\
% CD14	&		\\
% CD16	&		\\
% CD20	&		\\
% CD3	&		\\
% CD4	&		\\
% CD45	&		\\
% CD45RO	&		\\
% CD68	&		\\
% CD8a	&		\\
% E-Cadherin	& Tumour Suppressant		\\
% FOXP3	&		\\
% H3	&		\\
% IL-17a	&		\\
% Ki-67	&		\\
% LMNB1	&		\\
% Pan-Keratin	& Epithelial tissue marker		\\
% TCRgd	&	protype T cell	\\
% aSMA	&		\\
%     \hline 
%     \end{tabular}
%     \label{tab:labels}
% \end{table}

\subsection{Fluorescent Microscopy}%\label{AA}
Each of the IMC multiplex image datasets have a corresponding microscope image. These fluorescent microscopy (FM) images are captured at a much higher resolution than the IMC data though provide only a single optical measurement of the sample. The FM data provide a useful reference for the IMC data, see bottom left image in Fig.~\ref{fig:IMC}.

\subsection{Big Data}%\label{AA}
For this dataset there are 45 different IMC images, one per patient, ranging from 679,770 to 5,866,652 pixels. The median number of pixels is 2,488,800 with upper and lower quartiles of 3,403,338 and 1,988,820 pixels respectively. Each of these pixels have 19 labeled measurement tags and each IMC image has a corresponding FM image. This represents a large and problematic dataset to analyse. Therefore, in order to analyse these data in an efficient and practical manner, we train our models on a single dataset (which contains $\approx$10$^6$ pixels) and apply this to the remaining 44 patients. The end goal is to have a pre-trained model that clinicians can use with these data to aid rapid diagnosis of patients.  

% training time - Elapsed time is 316.201419 seconds.
% takes ~4s to read in a tiff stack 

\section{Methods}

\subsection{Deep Sparse Autoencoder}
Autoencoders are a special class of neural networks with a symmetric structure that encode data $\mathbf{X} \in \mathbb{R}^D$ into a different dimensionality space, $\mathbf{Z}\in \mathbb{R}^d$, before decoding this back to the original data space, $\mathbf{X}'\in \mathbb{R}^D$. Typically the network is constructed such that the original dimensionality $D$ is embedded into a lower dimensional space $d$. As the %symmetric property
output of the network is a reconstruction of the input,
the network can be trained using a loss function to minimise the difference between the input and decoded data and thus is an unsupervised method. The difference is typically formulated as a mean squared error,% $\epsilon$, %of the $M$ features and $N$ instances in the training set, as 
\begin{equation}
\label{eq:MSE}
    %\epsilon = \frac{1}{N} \sum_j^N \sum_i^M (x_{ji}-x'_{ji})^2 ~,
    %\epsilon = 
    \frac{1}{N} ||\mathbf{X} - \mathbf{X}'||^2 ~.
\end{equation}

%The encoded data, $z$, is obtained from via a weight matrix $\mathbf{W}$, bias vector $\mathbf{b}$ and activation function $\sigma$. The decoded data is obtained via similar operations  
The embedded space $z$ is obtained using an encoding transformation $E$, where as the reconstructed data $x'$ is recovered using a decoding transformation $D$
\begin{equation}
\label{eq:encode}
%{z} =\sigma(\mathbf{W} {x}+\mathbf{b} )~; \quad
%{x'} =\sigma'(\mathbf {W'} {z} +\mathbf{b}')~. 
z = E(x,\theta)~; \quad
x' = D(z,\theta')~,
\end{equation}
where $x \in \mathbf{X}$, $x' \in \mathbf{X}'$ and $z \in \mathbf{Z}$.
The parameters for the transformations, $\theta$ and $\theta'$, are optimised without labels by solving Eq.~(\ref{eq:MSE}) using scaled conjugate gradient descent \cite{moller_scaled_1993}. We select a sigmoid activation for both $E$ and $D$ transformations, due to its ability to capture nonlinear patterns in the data  \cite{LeCun2015,Thomas2016ssci,Thomas2019CMR}.

This canonical form can be extended in a number of ways, such as stacking several autoencoders where the encoded data from one layer becomes the input layer in the next layer. This yields a simple and fast method to train a deep autoencoder where each layer can be examined, or retrained, if desired. 

We employ regularisation terms to penalise non sparse solutions as these have been used in similar problems \cite{Thomas2017ssci}. Firstly, we employ a Tikhonov (L$_2$) regularisation term, $\Omega_\theta$, on $\theta$ and $\theta'$, to prevent over fitting in training and to reduce their complexity \cite{Krogh1992, Hassoun1995, McLoone2001}. This is computed for the $l^{th}$ layer as % This is calculated for $L$ hidden layers, $I$ inputs and $J$ output nodes,
\begin{eqnarray}
\label{eq:weight}
\Omega_\theta = \frac{1}{2}\sum_i\sum_j \left( \theta_{ij}^{(l)}\right)^2~.
\end{eqnarray}
Secondly, we include a sparsity penalty for neurons with a high activity, $\Omega_\rho$, enabling them to respond to specific features in the data. This is achieved by minimising the Kullback-Leibler divergence between a target level of activation, $\gamma$, and the average output $\hat{\gamma}$
\begin{eqnarray}
\label{eq:sparse}
\Omega_\gamma = KL(\gamma||\hat{\gamma})~,
%\sum_{k=1}^{K} \rho \log \left(\frac{\rho}{\hat{\rho_k}}\right) + (1-\rho)\log\left( \frac{1-\rho}{1-\hat{\rho}_k}\right) ~.
\end{eqnarray} 
The full cost function, $F$, to be optimised by the deep sparse autoencoder (DSA) is given by combining Eq.~(\ref{eq:MSE})-(\ref{eq:sparse})
\begin{eqnarray}
\label{eq:costFunction}
%F = \epsilon + \alpha \Omega_\theta + \beta \Omega_\gamma ~\\
F = \frac{1}{N} ||\mathbf{X} - D(E(\mathbf{X},\theta),\theta')||^2 + \alpha \Omega_\theta + \beta \Omega_\gamma ~,
\end{eqnarray} 
where $\alpha$ and $\beta$ are coefficients for the $\Omega$ regularisation terms.

A significant advantage of DSA is that once trained the encoding and decoding transformations are  deterministic. This has two direct benefits: 1) it enables training on a subset of the data allowing application to datasets that exceed the available computational resources available in terms of memory and speed; 2) pre-trained models can be transferred to other datasets providing a very efficient framework to analyse data from large experiments e.g. cohort and longitudinal studies. An example of the embedding space obtained from the DSA is given in Fig.~\ref{fig:scatterDensity}.

\subsection{Pseudo RGB Image}
\begin{figure}[t]
    \centering
    \includegraphics[width=0.45\textwidth]{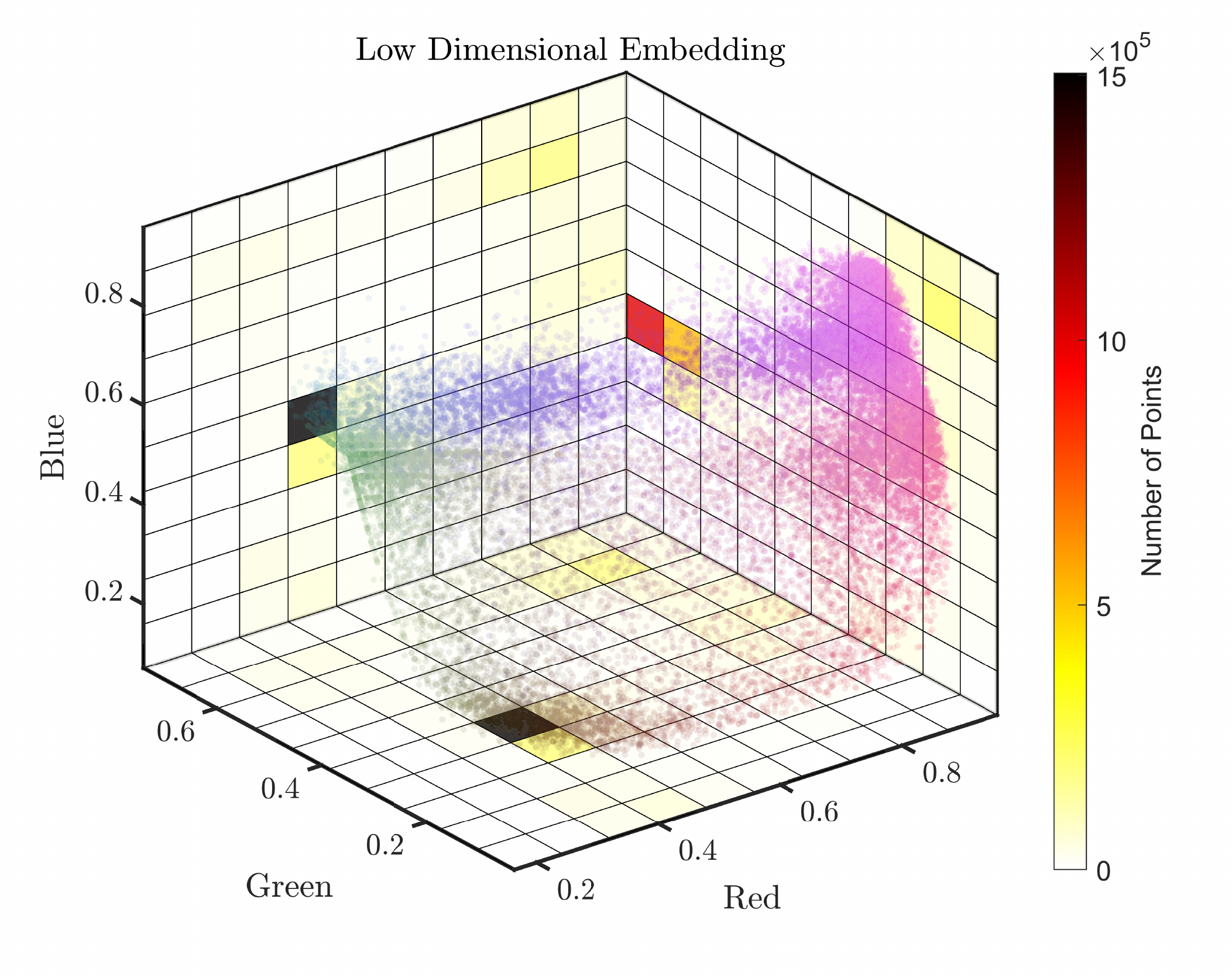}
    \caption{Deep embedded space from the DSA one of the IMC datasets. Heatmaps of the embedded density are also given for each of the pseudo RGB planes. Note the region of high density here corresponds to the background (non-tissue part) of the image. }
    \label{fig:scatterDensity}
\end{figure}

Here we use a DSA to embed the multiplex IMC image data into three dimensions in the so called bottleneck layer, also referred to as the latent space. This enables the visualisation of the data as a 3D scatter plot to identify patterns and differences. Additionally to treat each of the embedded dimensions as a pseudo RGB (red green blue) space providing a powerful summary of the multidimensional image in a single figure (Fig.~\ref{fig:scatterDensity}). Viewing the embedded data is beneficial over multicoloured overlays as it scales to any dimensional data (\cite{Thomas2016ssci, Thomas2017ssci} used this method for 1,000s of channels), avoiding the user from selecting a specific combination of images and colours. Moreover, the embedded space distinguish areas in the images of unique or multiple feature presence in a data driven way. That is, for components A and B, the DSA will distinguish between areas unique to A and B respectively, from those that contain combinations of both. Furthermore, for continuous data, it can also distinguish between different ratios of these components when both present, which may be particularly important in biological data. Obtaining such information in a data driven approach is highly favourable as the dimensionality of the data grows. 

\begin{figure*}[h]
    \centering
         \includegraphics[width=0.95\textwidth]{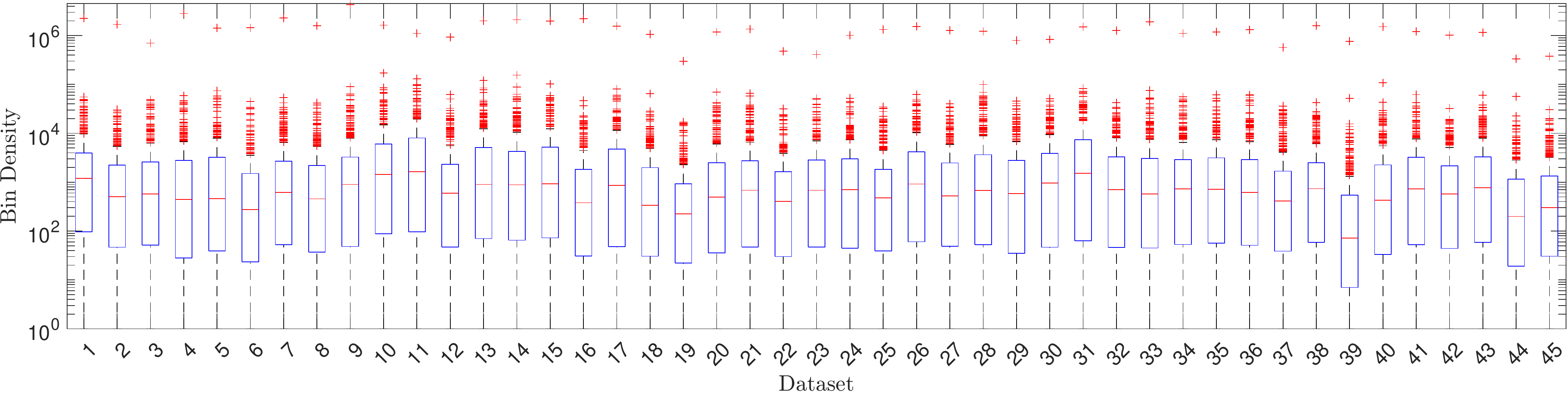}
        \caption{Boxplots of bin density in the embedded space for each 45 IMC datasets with outliers represented as red +. Note each dataset has a single high density outlier corresponding to the non-tissue region.  }
         \label{fig:boxplots}
\end{figure*}

\subsection{Density Based Estimation of Cluster Number}
Despite its ability to embed any dimensional data into an arbitrary dimensional space, which can reveal patterns and clusters, DSAs do not segment the data directly. It is possible to cluster the embedded space using methods such as $k$-means or DBSCAN, however this requires the selection of parameters for the number of cluster, or minimum points and radius respectively. With only a single parameter to tune, that may also be more intuitive to the data, we use $k$-means to cluster the embedded space and develop a data driven approach to determine the number of clusters based on the embedded density outliers. 

Selecting a sigmoid activation function in our autoencoder, 
\begin{equation}
    \sigma(x) = \left(  1+ e^{  - x  }   \right) ^{-1},~
\end{equation}
restricts the values of $z$ to the range [0 1] in each of the embedded dimensions. This property can be exploited as any arbitrary input data is embedded into a range [0 1] and we can therefore divide the embedding space into $B$ number of bins that are of fixed width and applicable to all input data via the DSA. For computational efficiency, as the total number of bins scales with $B^d$, where $d$ is the dimension of the embedded space, we define $B$=10 in all dimensions ($d$=3 for pseudo RGB) resulting in 1,000 bins across the embedded space. We can now count the number of points in each bin in the embedded space using the indicator function $\mathbf{1}_A$, 
\begin{equation}
    \mathbf{1}_A(x) := \begin{cases}
    1 \quad , & \text{if } x \in A\\
    0 \quad , & \text{if } x \notin A~,
    \end{cases}
\end{equation}
% change to the notation as hard to follow and looks ugly 
summing for all $N$ points in the data and for each bin $b$ in the embedded space. 
%That is we compute the bivariate histograms in each plane of the embedded space, in our case this is the $\emph{pesudo}$ $RG$, $RB$ and $GB$ planes of the embedding space. 
We count the number of points in bin $b$, with a width of $1/B$ as 
\begin{equation}
    \label{eq:hist}
%    \hat{f}^p_b(x) = \sum_{i=1}^N \mathbf{1}_{x_i \in b}(x_i).
 %   \hat{f}_b(x) = \sum_{i=1}^N \mathbf{1}_{b}(x_i).
    \eta_b(x) = \sum_{i=1}^N \mathbf{1}_{b}(x_i).
\end{equation}
We interpret the counts per bin as an estimation of point density in the embedded space and use this to determine the number of clusters in the embedded space, and therefore the input data. An example of the bin density of the embedded space is shown in Fig.~\ref{fig:scatterDensity}. 

We assume that the embedded space contains some dense regions and that there data are not homogeneously distributed in the embedding space, which we know from our prior knowledge of the data. The is, we know that biologically similar regions have similar chemical profiles and similar profiles will be grouped together in the embedded space to form dense regions. This has been observed in a number of studies that use dimensionality reduction of mass spectrometry data \cite{dexter_training_2020, abdelmoula_data-driven_2016,murta_implications_2021,smets_evaluation_2019,verbeeck_unsupervised_nodate}. %\textcolor{red}{Due to the resolution of the instruments compared to tissue features, there are typically a large number of pixels for a given feature. Biological and measurementz variability leads to distribution of values but these are far apart }
In the case where the data are homogeneous, clustering is not possible and a test for homogeneity can detect this automatically. Additionally we make the assumption that the number of dense regions is small compared to the number of bins, $\rho_n << B^d$, and for a large $B^d$ the majority of the bins should contain zero or a small number of points (e.g. noise data, outliers, etc). Therefore the distribution of $\eta_b(x)$ has a positive (right) skew with the dense regions as extreme points to the right (this is seen in Fig.~\ref{fig:boxplots}). This allows us to detect the dense regions automatically by computing the outliers of $\eta_b(x)$. The $b$th bin is considered an outlier if $\eta_b(x)$ is more than 1.5 times the inter-quartile range above the upper quartile. We then remove the bins below the 20$^{th}$ percentile to ensure we only select high density bins and avoid a region spanning several bins being counted as multiple clusters. 
% three scaled median absolute deviations from the median, 
% \begin{equation}
%     \xi_b(\eta_b) := \begin{cases}
%     1 \quad , & \text{if } |\eta_b - \tilde{\eta}| > 3\frac{-1}{\sqrt{2}\Psi(3/2)} \widetilde{|\eta_b - \tilde{\eta}|}      \\
%     0 \quad , & \text{otherwise } ~.
%     \end{cases}
% \end{equation}
% Here $\tilde{\eta}$ is the median of $\eta$ and $\Psi$ is the inverse complementary error function.
This provides a data driven estimator of the number for the clusters automatically from the embedded space that we can use for clustering algorithms such as $k$-means. 
% overall workflow figure? 

\begin{figure*}[t]
    \centering
    \includegraphics[width=0.95\textwidth]{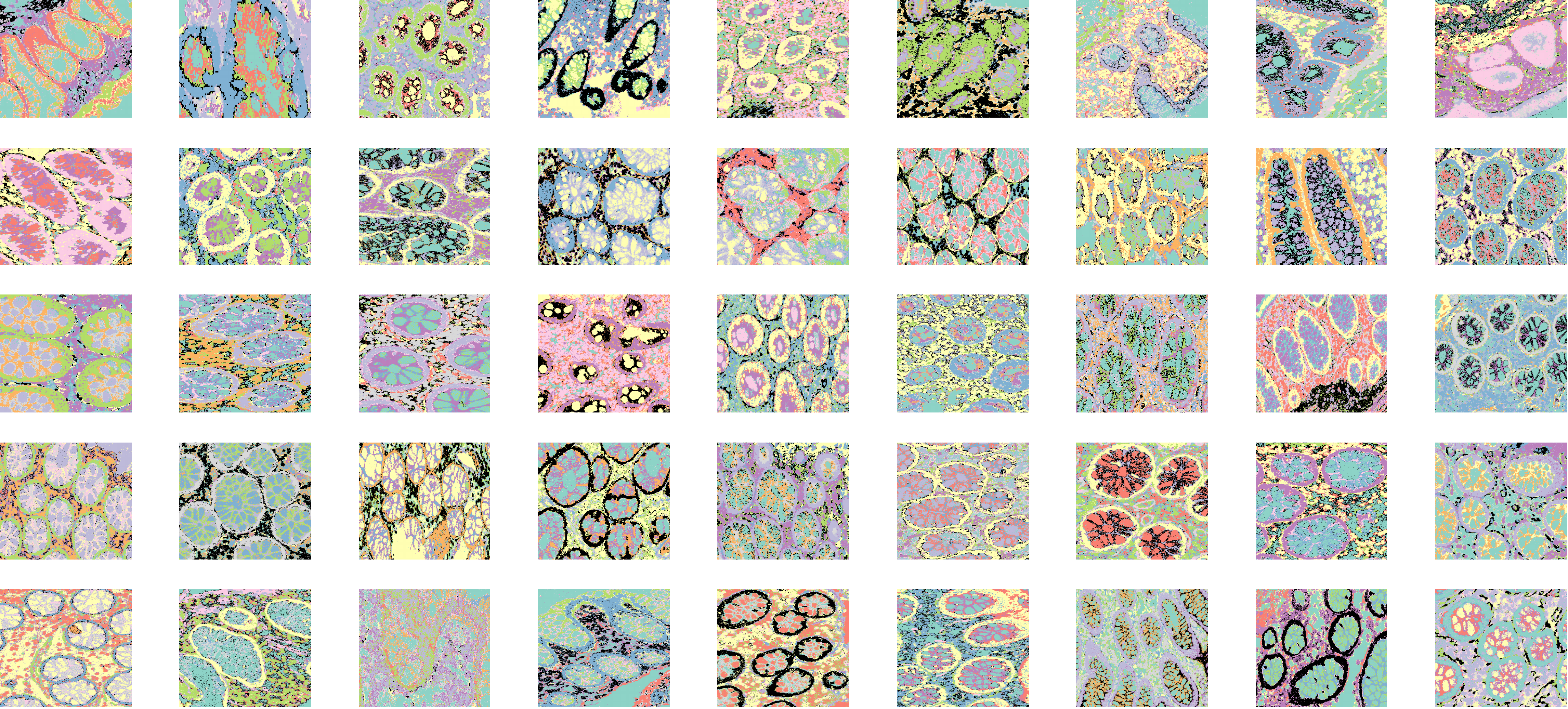}
    \caption{Clustering results for all 45 IMC images in this work. Images are obtained by $k$-means clustering of the embedded space using our embedded density outlier estimator for $k$. Note $k$ is estimated for each dataset individually and the colours are not comparable between images. Images are zoomed in for clarity and clearly showing sub cellular structures. }
    % grainy images high k?
    % add boxes to image? 
    %
%    \caption{Example images from the proposed work. Columns represent different tissues (reduced to 10 for clarity). the top row is the results from the deep autoencoder allowing visualisation of all IMC channels in a single pseudo-RGB image; the second row consists of the k-means clustering results from the density based data driven clustering of the embedded space from the autoencoder; the bottom row shows the FM images for reference. Due to the number of pixels in these images we provide a region of interest to show specific features. Note the estimation of $k$ from our method is provided in the top left corner of the clustered images in the middle row. The IMC image in the first column is used to train the deep autoencoder. }
    \label{fig:results10}
\end{figure*}

\section{Results and Discussion}
\label{sec:results}
We train the DSA on one of the IBD datasets and apply this encoding to the remaining 44 images, none of which are used in the training stage. We tune the parameters of the deep autoencoder during a preliminary set of experiments, though as in \cite{Cooke2022} we find a range of parameters suitable for this task. Here we use a network consisting of layers with 15, 10 and 3 hidden neurons respectively each trained for a maximum of 10,000 epochs with $\alpha$ = 10$^{-4}$, $\beta$=100 and $\hat{\gamma}$=0.5 for all layers. The total training time is 5 minutes 16 seconds using an NVIDIA GeForce RTX 3090 graphics card. 

As the 45 IMC images are from different patients there is no way to meaningfully average the clustering results across the datasets. Due to biological and technical variation, different patients as well as biopsies, may contain different tissue features so a universal $k$ for all images may not exist. Moreover, as we lack a expert annotations and there is some noise in the IMC data, we do not have a ground truth for comparison. Instead our method provides a estimate of $k$ for each dataset which we show is reasonable based on cluster correlation to IMC feature maps, silhouette index, estimation of $k$ based on the inflection point of the total sum squared distances plots, and visual inspection of the data in this section. 

We present the results of our method for all the 45 IMC datasets with summary metrics and graphs. The embedding of a single dataset by the DSA is given in Fig.~\ref{fig:scatterDensity} that also includes embedded density (i.e. Eq.~(\ref{eq:hist})) as heatmaps for each plane in the low dimensional space. This shows clear regions of high density in the embedded data that we use to estimate the number of features for clustering the data via $k$-means. These heatmaps clearly demonstrate that a large number of bins contain few or no points supporting the assumption that the number of dense regions is small compared to the number of bins, $\rho_n << B^d$.

The DSA can capture highly detailed features in the data in just three dimensions allowing for powerful visualisations of the multiplex data. The IMC data, despite a comparatively lower spatial resolution, contains richer information than the FM images, see Fig.~\ref{fig:IMC}. A single embedded image can provide a convenient overview of all these features that can be easily reviewed with the FM images, and may be more effective than making multiple comparisons wit the individual feature maps, particularly when the dimensionality of the hyperspectral image is large. The perceptual similarity of some colours can make distinction of regions challenging in the embedded space, however the clustered images can over come this challenge by clearly segmenting these regions (see Fig.~\ref{fig:results10} for instances from each IMC dataset). Clear sub cellular structures are visible in all 45 images indicating this is a robust method for analysing and segmenting these data. Each image in Fig.~\ref{fig:results10} depicts several cells with the same segmentation (within a given IMC dataset) of sub cellular components and other features. The zoomed regions in Fig.~\ref{fig:results10} are for clarity and visually demonstrate our methods effective estimation of $k$ for each dataset.

% \begin{figure*}[t]   
%     \centering
%     %  \begin{subfigure}[b]{0.95\textwidth}
%     %      \centering
%     %      \includegraphics[width=\textwidth]{figures/features_IBD_10.pdf}
%     %      %\caption{}
%     %      \label{fig:featuresMaps}
%     %  \end{subfigure}
%     %  \\
%     %  \centering
%      \begin{subfigure}[b]{0.95\textwidth}
%          \centering
%          \includegraphics[width=\textwidth]{figures/Correlation_IBD_10.pdf}
%          %\caption{}
%          \label{fig:correlation}
%      \end{subfigure}
%      \\
%      \begin{subfigure}[b]{0.89\textwidth}
%          \centering
%          \includegraphics[width=\textwidth]{figures/clusterMaps_IBD_10.pdf}
%          %\caption{}
%          \label{fig:clusterMaps}
%      \end{subfigure}
%         \caption{
%         % Montage of the 19 feature maps in the IMC data (top) of the data in the far right column of Fig.~\ref{fig:results10}. The specific components in each map is listed in Section~\ref{sec:IMC}.
%         (top) Correlation of each feature map (see Fig.~\ref{fig:IMC}) with (bottom) the corresponding cluster maps obtained from our method. %Note that all the IMC feature maps in the top row have saturated colour bars in order to aid coarse visualisation of the images.
%         \label{fig:correlationMaps}}
% \end{figure*}

\subsection{Clustered Features}
\begin{figure}[t]   
         \centering
         \includegraphics[width=0.45\textwidth]{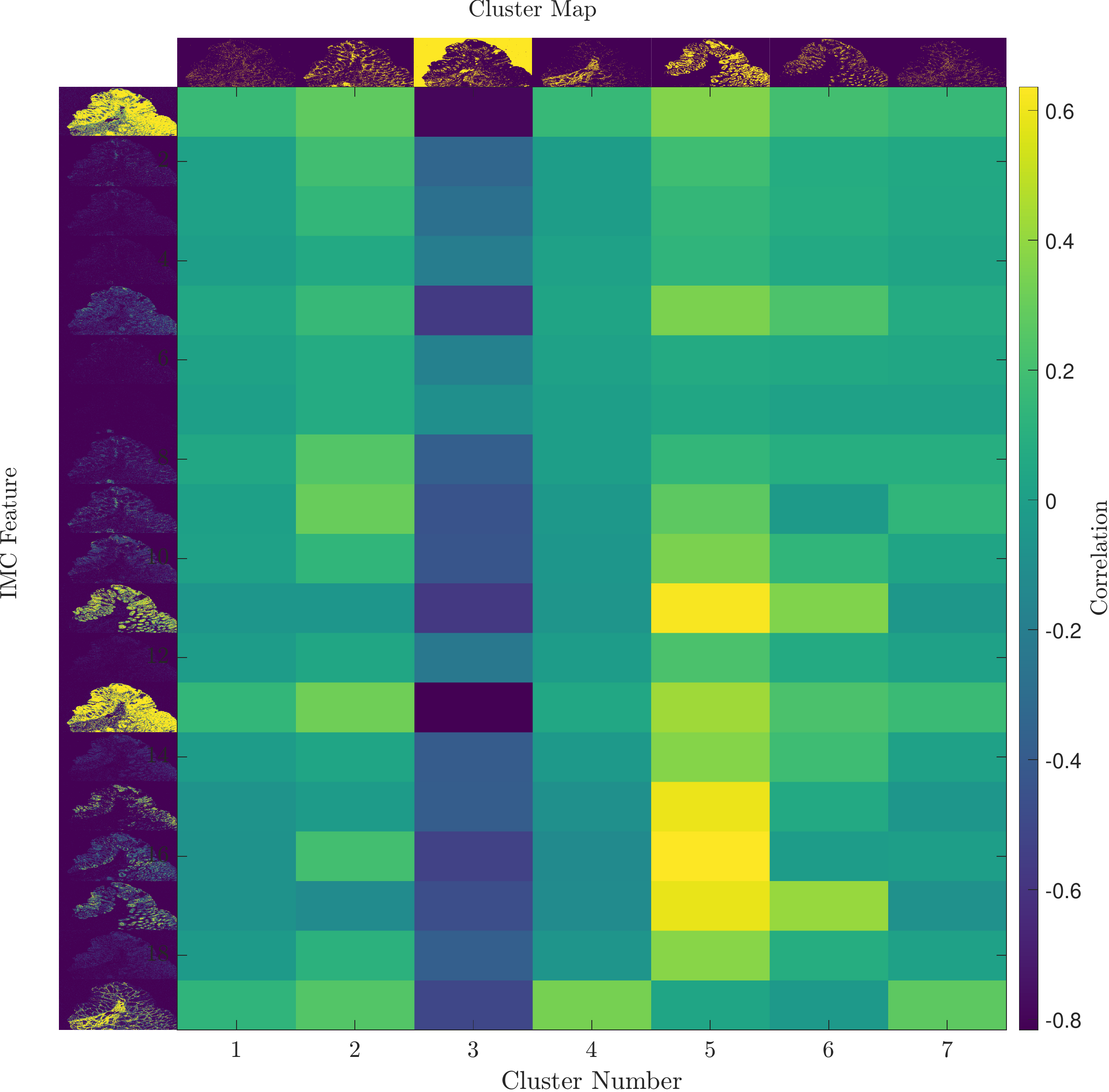}
        \caption{
        Correlation of each feature map (see Fig.~\ref{fig:IMC}) with the corresponding cluster maps obtained from our method. 
        \label{fig:correlationMaps}}
\end{figure}

%For brevity we examine the results of one of the 44 IMC datasets not used in training in detail, the right most column in Fig.~\ref{fig:results10}. 
%The embedded space for this dataset is shown in Fig.~\ref{fig:scatterDensity}. 
Firstly we consider the correlation of each of the cluster maps with the IMC feature maps as seen in Fig.~\ref{fig:correlationMaps}, for brevity we restrict this to just the dataset used to train the DSA. To compute the correlation with the binary cluster maps, we convert the IMC features maps to binary by considering any nonzero signal as 1. The colour scales for the feature maps in Fig.~\ref{fig:correlationMaps} have all been saturated in order to coarsely visualise the images for comparison with the cluster maps and reflect how the correlation was computed. This highlights the additional challenge in analysing IMC data of signal intensity and variation between feature maps that is overcome by our method. 

Several clusters correlate reasonably well with the IMC feature maps, though it is worth noting that clusters may incorporate information from several IMC features (due to the lower dimensionality) and hence a high correlation is not necessarily expected. This is highlighted by the fact that cluster 5 is correlated with several IMC feature maps and is visually similar to several feature maps (Fig.~\ref{fig:correlationMaps}).
Cluster 3 in Fig.~\ref{fig:correlationMaps}, corresponds to the non tissue region, is unsurprisingly anti-correlated with all feature maps, this also accounts for the highest density regions in the embedded space due to the large number of pixels and low variation in signal compared to the tissue regions.
Cluster 4 is visually similar to aSMA with a moderate correlation and uncorrelated with all others indicating that this feature is distinct from the rest. 
%The low correlation scores for the other clusters may be a result of the threshold used for IMC feature maps but are more likely due to the autoencoder generalising features. 
 %This is further indicated by the 2$^{nd}$ image in the 2$^{nd}$ row of Fig.~\ref{fig:correlationMaps} which shows two small regions that are present in a number of other feature maps (the next two images for example). These have been generalised into cluster 6. %This is actually captured in the deep autoencoder though is lost in the discretisation (clustering) of the continuous (embedded) space. 
% need
% LMNB1 
% aSMA
The features correlated with cluster 5 are 193Ir (DNA intercalator), H3 (chromatin/DNA marker), E-Cadherin (a tumour suppressant), Ki-67 (structural marker), LMNB1 (movement of molecules into/out of the nucleus), Pan-Keratin (an Epithelial tissue marker) and TCRgd (a protype T cell) \cite{baars_matisse_2021,ijsselsteijn_40-marker_2019}. Our method has grouped these features together in Cluster 5 indicating they are related functions or biological process.

In generalising to several feature maps, the correlation of a given cluster to each individual map may reduce. However, there is clear spatial information in the clusters that relates to the IMC feature maps. The observation that some clusters contain specific features while others generalise across several demonstrates flexibility of this method. The ability to identify general regions of co-location of features as well as isolating unique regions is important in biological and pharmaceutical applications where the presence and absence of specific chemicals is of vital importance. 

% https://www.frontiersin.org/articles/10.3389/fimmu.2019.02534/full
%Structural markers: Vimentin (purple), Ki-67 (green), D2-40 (red), and Keratin (cyan). 
%Myeloid markers: CD68 (red), CD163 (blue), HLA-DR (green), and CD11c (white). 
%Lymphoid markers: CD3 (red), CD8 (yellow), CD4 (blue), and FOXP3 (cyan).
%T-helper cells (A1, CD3+CD4+)
%regulatory T cells (A2, CD3+FOXP3+),
%cytotoxic T cells (A3, CD3+CD8+), 
% macrophages HLA-DR+ macrophages (B1, CD68+HLA-DR+), CD163+ macrophages (B2, CD68+CD163+), and CD11c+ macrophages (B3, CD68+CD11c+), 
% tissue resident cytotoxic T cells (C1, CD3+ CD8+CD103+), tissue resident T helper cells (C2, CD3+CD8−CD103+), and activated T cells (C3, CD3+GZMB+).

\subsection{Silhouette Index}
\begin{figure*}[t]      

    \begin{subfigure}[b]{0.9\textwidth} 
        \centering
        \includegraphics[width=\textwidth]{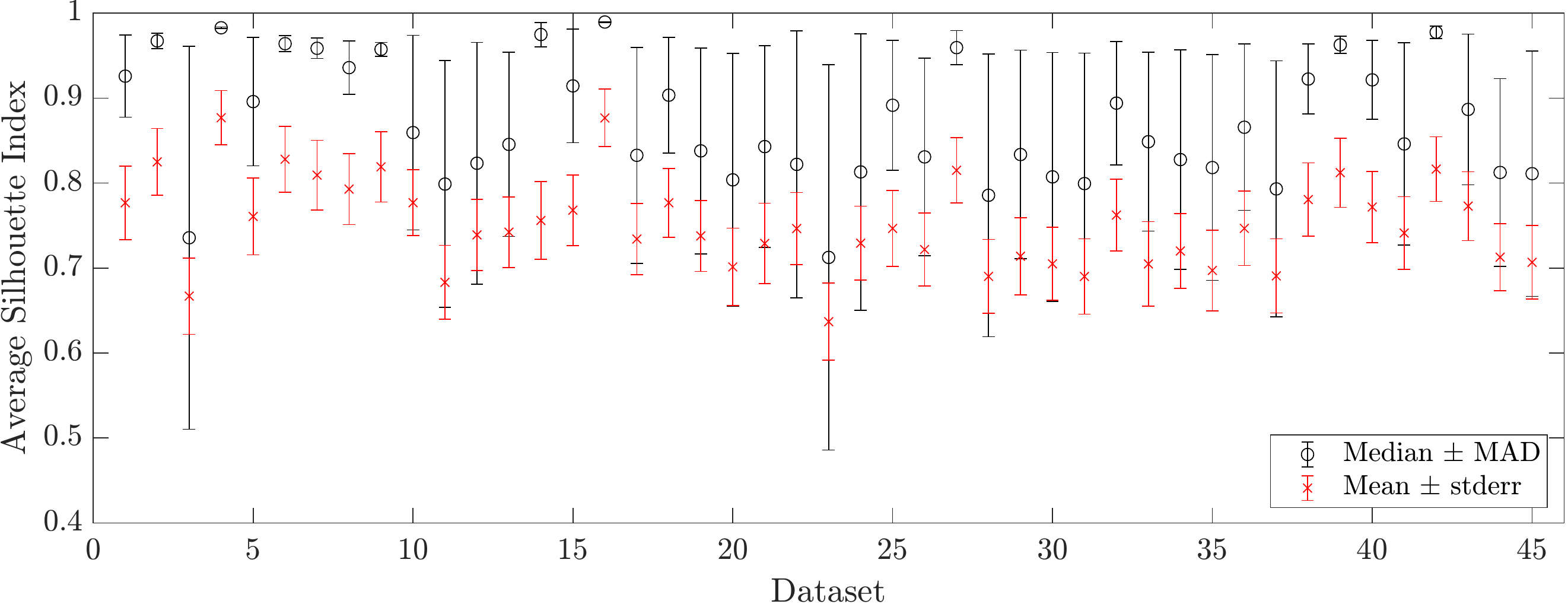}
        \caption{Average silhouette index for each dataset}
        \label{fig:silhouette}
    \end{subfigure}
    % \begin{subfigure}[b]{0.43\textwidth} 
    %     \centering
    %     \includegraphics[width=\textwidth]{figures/silhouette_IBD_10.pdf}
    %     \caption{}
    %     \label{fig:silhouette}
    % \end{subfigure}
    % \hspace{1.5em}
    %  \begin{subfigure}[b]{0.43\textwidth}
    %      \centering
    %      \includegraphics[width=\textwidth]{figures/ElbowPlot_every100points_scaled.pdf}
    %      \caption{}
    %      \label{fig:elbow}
    %  \end{subfigure}
     \\ ~\\ %\vspace{1.5em}
    %  \begin{subfigure}[b]{0.95\textwidth}
    %      \centering
    %      \includegraphics[width=\textwidth]{figures/DensityBoxplots.pdf}
    %      \caption{}
    %      \label{fig:boxplots}
    %  \end{subfigure}
    %  \\ \vspace{0.5em}
     \begin{subfigure}[b]{0.9\textwidth}
         \centering
         \includegraphics[width=\textwidth]{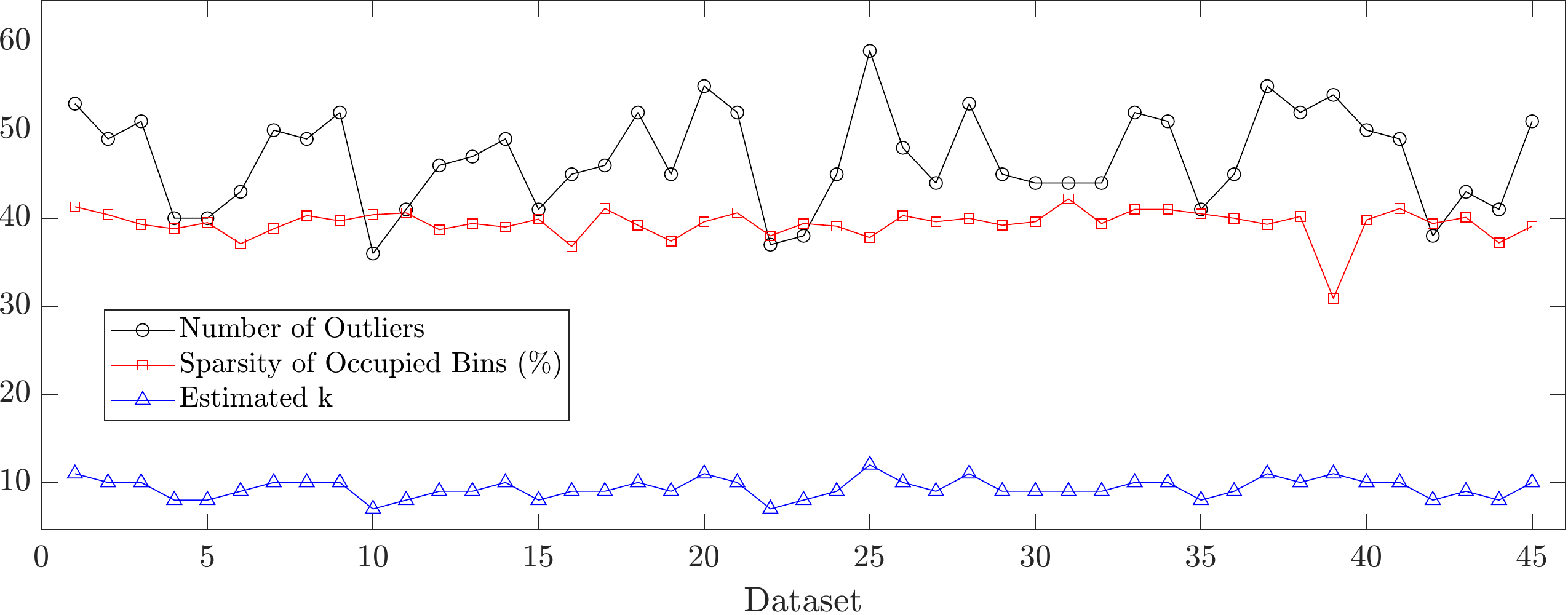}
         \caption{Comparison of all 45 IMC datasets}
         \label{fig:outliers}
     \end{subfigure}
        \caption{
        %(a) Boxplots of the silhouette index for each cluster in the training data.
        %(b) Elbow plot for a subset of pixels in all 45 IMC datasets.
        (a) Average silhouette index for each dataset. Averages are the median (black circles) $\pm$ the median absolute deviation (MAD), and mean (red crosses) $\pm$ the standard error in the mean (stderr).
        %(b) Boxplots of bin density in the embedded space for each 45 IMC datasets.
        (b) Estimated $k$, number of outliers and sparsity $\%$ for each IMC dataset.
        \label{fig:evaluation}
        }
\end{figure*}

The quality of the Clustering is shown via the silhouette index in Fig~\ref{fig:silhouette}. The high median values (all $>$0.71) indicate the number and membership of identified clusters is good and the majority of points have medians above 0.80. Some individual members of clusters (i.e. pixels) yield lower silhouette scores which may be due to noise in the data, the quality of the embedding, or the clustering process. The mean silhouette index is lower than the median in all cases but is $>$0.70 in most cases. The mean is likely to be less robust than the median given the skewness in these distributions. Robust clustering of experimental data is a challenging task where differences may be due to the underlying sample, acquisition method, noise, or data processing for example. Further investigation is required to determine this, though we do note that the high median scores indicate the embedded density method is a reasonable estimate of the number of clusters. 

\subsection{Embedding Unseen Data}
The trained DSA provides a consistent embedding space for all images providing a very efficient way to analyse a large number of images. As the clustering is done individually on each IMC dataset the number of clusters, and the assignment to specific regions varies with each dataset (see Fig.~\ref{fig:results10} and Fig.~\ref{fig:outliers}). Matching cluster regions between datasets is beyond the scope of this work, but could be achieved via linkage algorithms or performing t-tests for example. %The median number of clusters in the 45 IMC images is 12 with a range of 9 to 19. 

\subsection{Comparison of $k$ and Runtime}
\begin{figure}
    \centering
    \includegraphics[width=0.45\textwidth]{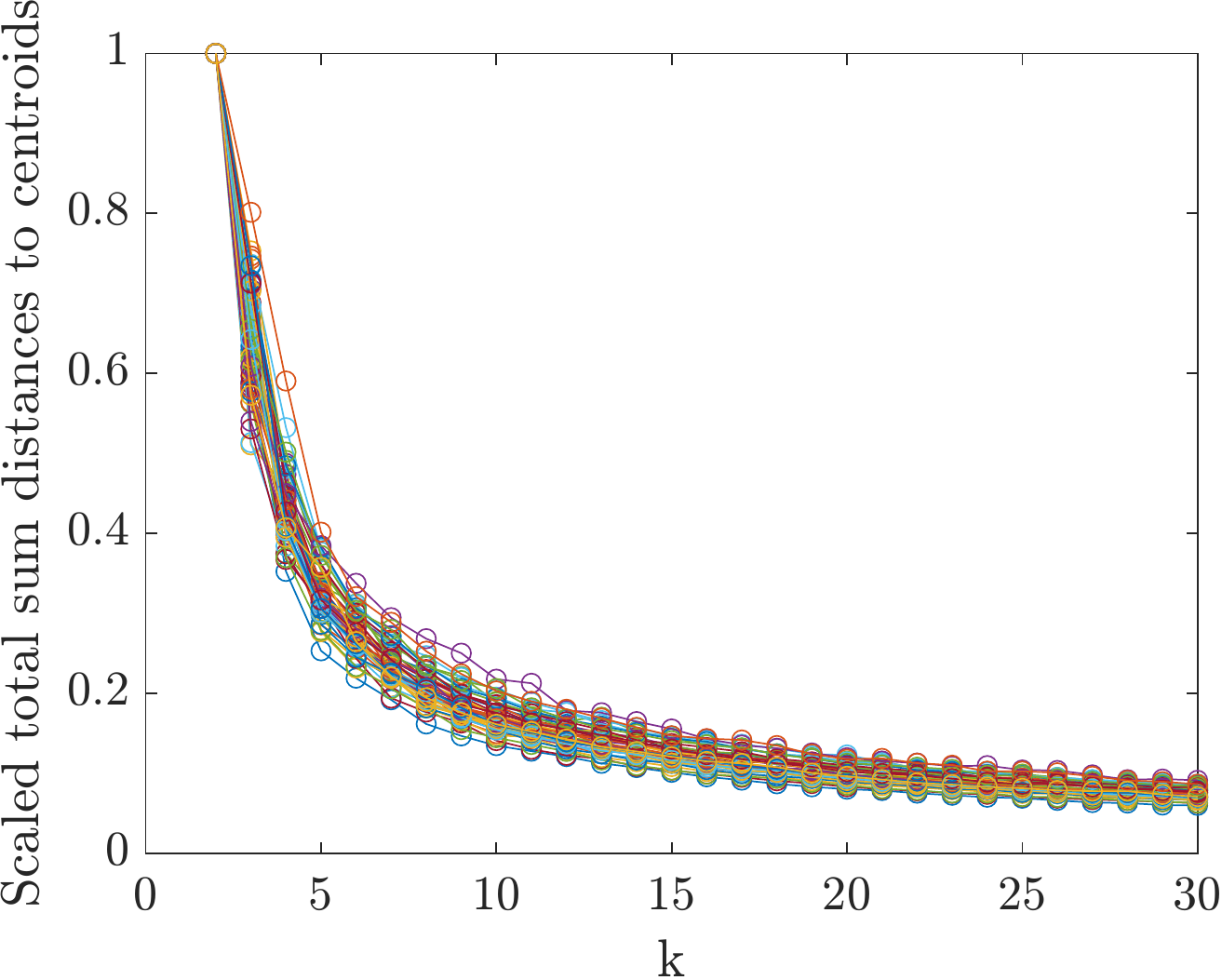}
    \caption{Total sum squared distance to cluster centroids scaled by the maximum values, as a function of cluster number $k$ for all 45 IMC datasets.}
    \label{fig:elbow}
\end{figure}

Another estimator of $k$ is to identify the inflection point in the total sum squared distances (SSD) to cluster centroids for increasing $k$ \cite{yuan_research_2019}. This however requires the data first to be clustered over a wide range of $k$ which is computationally restrictive for large data such as the IMC data here. Moreover, this is extremely expensive for a single dataset and in this work we analyse 45 making this an extremely expensive alternative. However, as the inflection point estimator provides a comparison to our method we calculate the SSD for increasing $k$ for all 45 IMC dataset, using a subset of each dataset to reduce the computational burden. Scaling the SSD by the maximum allows all 45 curves to be clearly viewed together revealing that they all show a plateau around $k\geq$20 (Fig.~\ref{fig:elbow}). The average $k$ estimated across the IMC datasets is given in Table~\ref{tab:k_estimate}. Due to a combination of noise and sub-setting the data, the SSD plots exhibit fluctuations and determining the exact inflection point is not robust. Hence we also determine the inflection point within a small tolerance (0.005). Our method compares well to both inflection point methods and only requires a single clustering run per dataset, compared to the 30 needed to generate the SSD plot in Fig.~\ref{fig:elbow}. %This constitutes to a significant computational saving as we will outline in Section~\ref{sec:runtime}.

\begin{table}[h]
    \centering
    \caption{Estimated $k$ averaged over the 45 IMC datasets with mean ($\pm$ standard error in the mean) and median ($\pm$ median absolute deviation). The Inflection Point is computed when the second derivative equals zero or when this is within a small tolerance factor (Tolerance) set to 0.005. }
    \begin{tabular}{l|c|c}
        \hline
        $k$ Estimation & Mean $k$ & Median $k$ \\
        \hline
        Inflection Point & 12.67 $\pm$ 0.50 & 13 $\pm$ 2 \\
        Inflection Point (Tolerance) & 8.69 $\pm$ 0.19 & 9 $\pm$ 1 \\
        Density Outlier & 9.11 $\pm$ 0.16 & 10 $\pm$ 0 \\
        \hline
    \end{tabular}
    \label{tab:k_estimate}
\end{table}

%\subsection{Runtime}
%\label{sec:runtime}
Furthermore, the estimation of $k$ prior to clustering is very efficient compared to the SSD method, estimating $k$ in $\approx$2~s and $\approx$700~s respectively per dataset (see Table~\ref{tab:runtimes}). This equates to 1.5~minutes for our methods and 8.6~hours for the SSD method for all 45 IMC images. The training time of the autoencoder was 316 s for the training data and is only required once for the 45 IMC datasets. This is less than half the time to generate an SSD plot for a single IMC dataset, highlighting the significant computational savings with this method. 

% with a mean time of 0.21 $\pm$ 0.01 seconds per IMC dataset. The embedding time using the pre-trained autoencoder takes an average of 1.83 $\pm$ 0.10 seconds per dataset yielding an overhead of only 2 seconds to determine an estimate of $k$ before clustering compared to generating an elbow plot, which takes an average of %7.89 $\pm$ 0.53 seconds per dataset when using only 2\% of data points due to computational resource limitations. 
% 691.67 $\pm$ 43.61 seconds per dataset, 

\begin{table}[h]
    \centering
    \caption{Runtimes for estimating $k$. $^*$The autoencoder training time is averaged over the 45 datasets. }
    \begin{tabular}{l|c|c}
        \hline
        Method & per dataset (s) & all datasets (s)   \\
        \hline
        Autoencoder Training$^\dagger$ & 316.01 & 316.01 \\
        Embedding with autoencoder & 1.83 $\pm$ 0.10 & 82.35 \\
        Density Estimation & 0.21 $\pm$ 0.01 & 9.45 \\
        Outlier Detection & 1.78 $\pm$ 0.33 $\times$ 10$^{-4}$ & 0.01 \\
        Total & 2.04  & 91.81 \\
        Total (inc $^\dagger$) & 9.06$^*$  & 407.82 \\
        \hline
        SSD plot inflection & 691.67 $\pm$ 43.61 & 31125.15 \\
        \hline
    \end{tabular}
    \label{tab:runtimes}
\end{table}

%For the sake of brevity we only include detailed analysis for one of the IMC datasets. However we include a montage image of all the embedded IMC data sets in Fig.~\ref{fig:montage} to show the consistency in the embedded data used for clustering. 
% \begin{figure*}[!t]
%     \centering
%     \includegraphics[width=0.95\textwidth]{figures/IBD_all_45_low.png}
%     \caption{Montage of deep autoencoder embedding for the 45 IMC datasets visualised as pseudo RGB images and showing highly consistent representations of the data. }
%     \label{fig:montage}
% \end{figure*}

\section{Conclusion}
In this work we have developed a method for estimating the number of clusters in order to analyse multiplex cell imaging data. By using a unsupervised method to embed the data into a fixed range space, our method can determine the cluster number automatically via the density of regions in the space using the assumption that there are far few clusters than binned regions. We have demonstrated the use of this in detail with a single IMC dataset of colonic tissue from a patient with irritable bowel disease. Moreover, we show how this can be readily applied to other dataset as a efficient means to analyse large sets of multiplex images, 45 IMC images in this work. Our method is purely data driven and is able to circumvent the need for selecting a number of cluster {\emph a priori} thus providing a powerful approach for unguided data exploration. Finally, our method is extremely efficient and we show a reduction in computation time by two orders of magnitude compared to estimating via the total sum squared distances as a function of cluster number. Our generic methodology can be applied to other types of multiplex and hyperspectral data as well as utilising any future developments for embedding high dimensional data by replacing the DSA stage. Further developments of the method could include the usage of other clustering methods, e.g hierarchical clustering for consistent clustering results. 

%\section*{Acknowledgment}

%\section*{References}

\bibliographystyle{unsrt2authabbrvpp}
\bibliography{refs,CIBCB2022refs, DLCMR}

\end{document}